\documentclass{article}

\usepackage[final,sglblindworkshop]{neurips_2026}
\workshoptitle{Evaluation of Interactive Agents}

\usepackage[utf8]{inputenc}
\usepackage[T1]{fontenc}
\usepackage{hyperref}
\usepackage{url}
\usepackage{booktabs}
\usepackage{amsfonts}
\usepackage{nicefrac}
\usepackage{microtype}
\usepackage{xcolor}
\usepackage{graphicx}
\usepackage{float}
\usepackage{placeins}
\usepackage{multirow}
\usepackage{amsmath}
\usepackage[font=small]{caption}

\newcommand{\luca}{\textsc{Luca}}

\title{\luca: A Synthetic World for Experimental Reasoning}

\author{%
  Anna Gogleva\thanks{Lead and corresponding author.} \quad Thomas Charman \quad Martin Balla \quad Karl Gemayel \\
  Relation Therapeutics \\
  \texttt{anna.gogleva@relationrx.com}%
}

\begin{document}
\raggedbottom

\maketitle

\begin{center}
\fbox{\begin{minipage}{0.92\textwidth}
\small
\textbf{Author note (November 2026).} This paper reports results from the
original \luca{} environment. Subsequent fixes address implementation issues
affecting some results and comparisons, so evaluations using the updated
repository may yield different performance numbers. Details are documented in
\texttt{KNOWN\_ISSUES.md}. We are developing \luca{} v2 and plan to report
updated evaluations in a future revision.
\end{minipage}}
\end{center}

\begin{abstract}

An experimental agent can succeed for the wrong reasons.  Evaluations, however, often focus on the outcome, not the experiments, evidence, and inferences that produced it. We introduce \luca{}, a configurable synthetic world for evaluating this process under fully known ground truth. Agents learn unknown regulatory systems through budgeted experimentation, then predict unseen interventions, diagnose hidden faults, and attempt rescues. This setup allows us to evaluate experiment selection, prediction, diagnosis, and intervention separately.

Across seven frontier LLMs and three baselines, these capabilities dissociate sharply. Agents frequently rescue malfunctioning systems while diagnosing the wrong fault; diagnosis varies 7× across models while prediction varies only 1.4×. Models also adopt surprisingly different experimental strategies, and greater exploration does not reliably yield better performance. In one failure mode, agents often perform the experiment needed to distinguish competing causes, but still fail to use the resulting evidence. \luca{} turns these hidden failures into measurable behaviours, providing a controlled environment to study where experimental agents succeed and fail.
Code: \url{https://github.com/RelationRx/luca}.
\end{abstract}

\section{Introduction}
\label{sec:intro}

Experiments allow us to understand how systems work. This requires deciding what to test, how to interpret the evidence and how to use it to make predictions, how to diagnose problems, or choose interventions~\citep{boiko2023autonomous,lu2026automation}. However, a successful outcome does not necessarily imply a sound understanding: an agent may find an effective intervention while holding an incomplete or incorrect view of the system.

Evaluating these stages separately is difficult with existing frameworks. Interactive-agent benchmarks (AgentBench~\citep{liu2024agentbench}, WebArena~\citep{zhou2024webarena}) score sequential action and feedback, discovery environments (DiscoveryWorld~\citep{jansen2024discoveryworld}, DiscoveryBench~\citep{majumder2025discoverybench}, ScienceAgentBench~\citep{chen2025scienceagentbench}) score what an agent achieves, and causal-discovery benchmarks (CausaLab~\citep{yang2026causalab}, CausalBench~\citep{chevalley2022causalbench}) score what it recovers about structure, with CausaLab also scoring a held-out prediction. Each of these approaches scores part of this process rather than the whole chain from experiment choice to inference to action. In real systems, the underlying mechanism and the relevant counterfactuals are rarely known well enough to score all of these stages directly.

We introduce \luca{}, named after the Last Universal Common Ancestor~\citep{weiss2016luca}, a configurable synthetic world for evaluating this experimental process. An LLM agent learns an unknown dynamical system by perturbing it under a fixed experimental budget and observing downstream readouts. The agent is then tested on unseen intervention outcomes. In malfunctioning systems, it also must diagnose a hidden fault before attempting a rescue. Because the complete system is known to the evaluator, prediction, diagnosis, rescue, and experimental strategy can be scored separately and deterministically. The commitment-before-rescue design further prevents agents from using a successful rescue to reverse-engineer their diagnosis.

Inspired by experimental biology, \luca{} instantiates its systems as synthetic regulatory networks, using gene regulatory structure as a template for a general synthetic world rather than attempting to simulate cellular biology realistically. Perturbations propagate through interacting variables, effects are nonlinear, components can be hidden, and compensatory interventions can succeed without identifying the original fault. Synthetic construction gives us direct control over these properties and complete access to the mechanism and its counterfactuals. All entity names are invented, so memorised biology cannot substitute for experimentation.

Across seven frontier LLMs, \luca{} reveals substantial reasoning differences hidden by task success. Agents frequently rescue malfunctioning systems while diagnosing the wrong fault. In a scaling study across four of these models, agents reach similar accuracy from amounts of evidence that differ by more than an order of magnitude.

\paragraph{Contributions.}
\begin{itemize}\setlength{\itemsep}{0pt}\setlength{\parskip}{0pt}
    \item \luca{}, a configurable synthetic environment that exposes the experimental trajectory and separately evaluates experimental strategy, prediction, diagnosis, and intervention.
    \item An evaluation of seven frontier LLM agents showing a systematic dissociation between successful intervention and correct diagnosis.
    \item An analysis of experimental strategy and scaling showing large behavioural differences between agents, including some that reach comparable accuracy from very different amounts of evidence.
\end{itemize}

\section{Benchmark Design}
\label{sec:benchmark}
\luca{} represents each unknown system as a synthetic organism that an agent can
probe through interventions and observations. The benchmark has two components:
the construction of these organisms and the interactive tasks through which
agents experiment with them.

\subsection{Synthetic Organism Design}
\label{sec:design}

\begin{figure}[!b]
\centering
\includegraphics[width=0.90\linewidth]{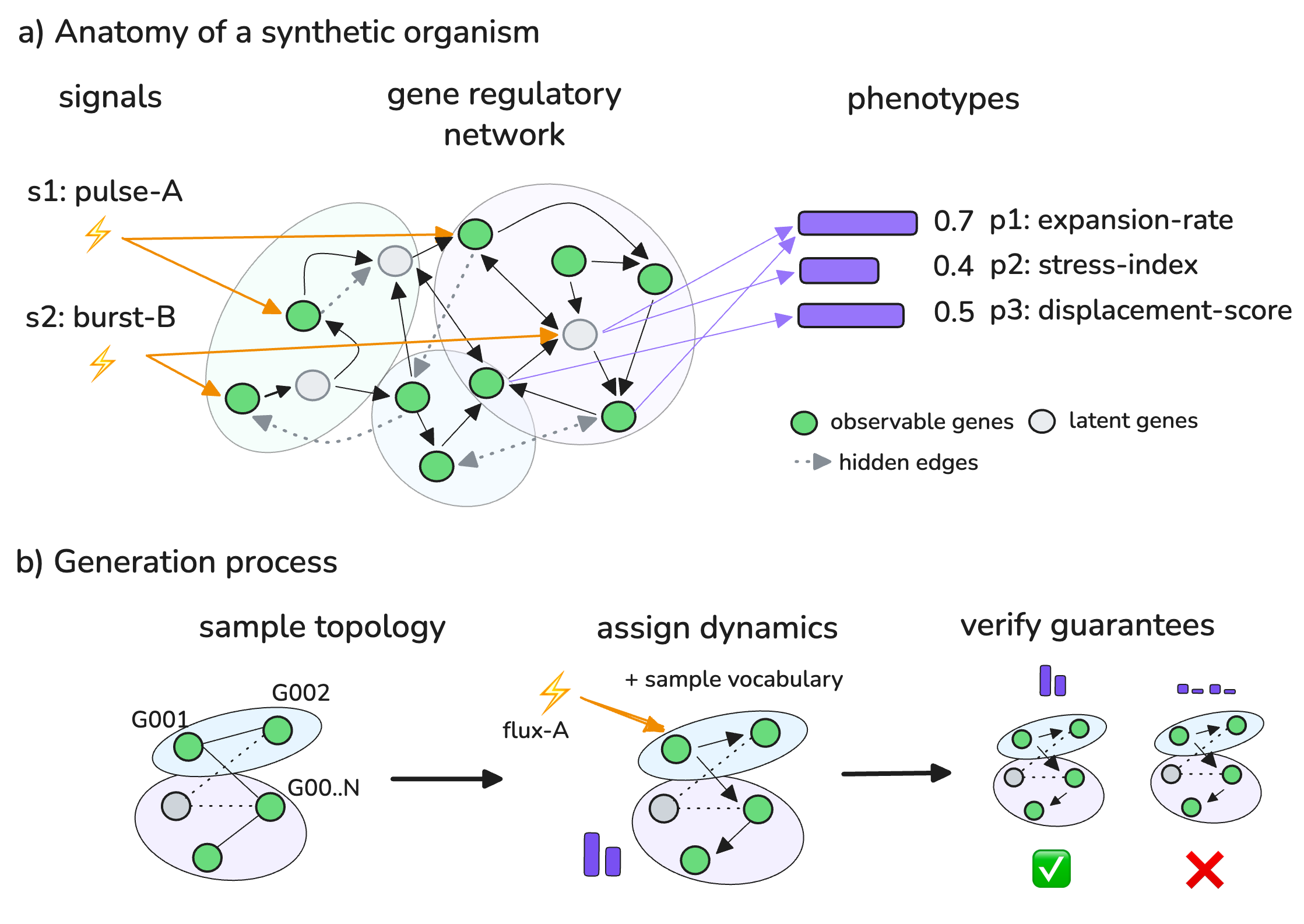}\vspace{-4pt}
\caption{\luca{} design. (a) Organism anatomy: signals, GRN, and
phenotype readouts at steady state. Large pale ellipses denote modules;
green nodes are observable
genes and grey nodes latent ones; black arrows are network edges, solid where
observable and dotted where hidden, while orange arrows carry signals and
purple arrows drive phenotype readouts.
(b) Generation pipeline; each organism is verified against construction
guarantees.}
\label{fig:organism}
\end{figure}

Each organism is built around a synthetic gene regulatory network (GRN) with $N$ genes
organised into $M$ functional modules. External signals can drive gene states directly.
Together, gene interactions and signals determine a set of observable \emph{phenotypes},
the high-level outputs of the organism.
Organisms can additionally be placed in a \emph{disease mode}, in which one gene
or signal is locked into an abnormal state and perturbs downstream phenotypes.
Figure~\ref{fig:organism} shows the anatomy and the generation process.

To simulate the organism, gene states evolve via a sigmoid update rule~\citep{mjolsness1991connectionist,karlebach2008modelling}:
\begin{equation}
  s_i \;\leftarrow\; \sigma\!\left(\beta\!\left(\sum_{j=1}^{N} w_{ij}\, s_j + c_i + b_i\right)\right) + \varepsilon_i,
  \qquad \sigma(x) = \tfrac{1}{1+e^{-x}}.
\end{equation}
Here $s_i \in [0,1]$ is the state of gene $i$, $w_{ij}$ are regulatory weights (positive for activation, negative for
inhibition), $b_i$ is a gene-specific bias, and $c_i$ is the summed contribution
of active external signals targeting gene $i$.
The steepness parameter $\beta$ controls how nonlinear the dynamics are,
and $\varepsilon_i \sim \mathcal{N}(0,\sigma^2)$ adds per-gene measurement noise
(see Table~\ref{tab:tiers}).
The network is iterated to a fixed point.

\paragraph{Invented vocabulary.}
All entity labels (gene identifiers, module names, signal names, phenotype names)
are drawn from a fixed lexicon of constructed words with no correspondence to real
biological entities, eliminating the retrieval confound.
For example, genes are labelled \texttt{G001}, \texttt{G002}, \ldots and phenotypes
appear as \texttt{stress-index} or \texttt{displacement-score}, constructed terms
absent from any biological database.

\paragraph{Difficulty tiers.}
Organisms are generated at three difficulty tiers (Table~\ref{tab:tiers}). We
vary edge density, feed-forward loops, feedback edges, sigmoid steepness,
and latent gene fraction. Easy organisms have near-linear dynamics with no
hidden variables; hard organisms have dense feedback and 25\% latent genes
acting as hidden confounders. All parameters scale with $N$.

\begin{table}[h!]
\centering
\caption{Difficulty tier definition.}
\label{tab:tiers}
\small
\begin{tabular}{lccc}
\toprule
 & Easy & Medium & Hard \\
\midrule
\multicolumn{4}{l}{\emph{Generative parameters (fixed per tier)}} \\
Observability              & 0.80  & 0.60  & 0.50 \\
Intra-module edge density  & 0.25  & 0.40  & 0.50 \\
Inter-module edge density  & 0.04  & 0.08  & 0.15 \\
Inhibition fraction        & 0.20  & 0.30  & 0.40 \\
Sigmoid steepness $\beta$  & 0.7   & 1.0   & 2.0 \\
Noise $\sigma$             & 0.005 & 0.010 & 0.030 \\
Latent gene fraction       & 0.0   & 0.0   & 0.25 \\
\midrule
\multicolumn{4}{l}{\emph{Emergent statistics at $N{=}20$ (vary per seed)}} \\
GRN edges                  & 26--42  & 58--75  & 87--121 \\
FFLs                       & $\sim$1 & $\sim$3 & $\sim$4 \\
Feedback edges             & 0       & $\sim$3 & $\sim$5 \\
\bottomrule
\end{tabular}
\end{table}

\paragraph{Construction guarantees.}
To prevent invalid organisms we enforce four properties: 1) every observable gene contributes to at least one phenotype readout, so no perturbation is invisible to the agent; 2) held-out gene combinations have non-additive joint effects, so predicting them requires a genuine causal model; 3) in hard-tier organisms, every latent gene shifts at least one phenotype, ensuring hidden genes are true confounders; 4) phenotype baselines are re-centred to 0.5 to ensure detectable perturbation shifts. Full generation details are in Appendix~\ref{app:generation}. Every organism is fully determined by a \texttt{(seed, tier, N)} triple, so any result is exactly reproducible without stored files.

\subsection{Task Design and Evaluation}
\label{sec:interface}

An agent interacts with an organism through a budget-limited API. In healthy mode, the session proceeds in two phases: exploration and prediction. Disease mode adds a third phase, where the agent must diagnose the malfunction and attempt a rescue intervention.

\paragraph{Exploration.}
The agent runs experiments by applying gene interventions
(\textsc{block}: suppress to zero; \textsc{boost}: saturate to max;
\textsc{dampen}: partial reduction) and observing the resulting
steady-state phenotypes.
The session budget is $B = N + N_{\text{sig}} + 8$, where $N_{\text{sig}}$ is the
number of external signals ($B{=}31$ at $N{=}20$).
This is sufficient for a complete single-gene sweep but leaves little room for
targeted follow-up, forcing strategic prioritisation.

\paragraph{Prediction tasks.}
After exploration, the agent receives two held-out two-gene prediction tasks
on the same organism. Each task is a simultaneous intervention on a pair of
observable genes drawn from the same pool the agent could perturb during
exploration. The joint outcome is never revealed during exploration.
Pairs are sampled to share a phenotype readout, so their joint effect is
non-additive due to sigmoid nonlinearity. Correct predictions therefore
require a genuine causal model of the organism, not simply combining
individual gene effects (Appendix~\ref{app:generation}).

\paragraph{Disease mode.}
In disease mode the organism has a \emph{malfunction}: one gene or signal is
locked into an abnormal state. This drives one or more phenotypes outside their
healthy reference ranges.
By construction, at least one corrective single-target intervention exists
that restores $\geq\!1$ aberrant phenotype to the healthy range.
Malfunction targets are never latent genes, ensuring the agent can act on a
correct diagnosis.
The agent must (1)~\textbf{diagnose} the malfunction by naming the gene and
malfunction type, and (2)~\textbf{rescue} by proposing a corrective
intervention.
The agent cannot test rescue interventions until it has committed to a
malfunction hypothesis. This commitment-before-test design is essential:
without it, agents could brute-force rescue and back out a diagnosis,
collapsing the two metrics into one.

\paragraph{Construction guarantees rule out shortcuts.}
Two-gene tasks share a phenotype readout, so combining single-gene responses
cannot recover the joint outcome. Invented gene names remove retrieval priors.
In disease mode, the diagnosis score is withheld until session end and rescue
is gated behind a committed diagnosis, blocking both backward reasoning and
rescue-to-diagnosis attacks.

\paragraph{Evaluation metrics.}
Table~\ref{tab:metrics} summarises all metrics. The composite score
is prediction accuracy (healthy) or the mean of prediction, diagnosis,
and rescue (disease).
We withhold the diagnosis score until session end, preventing backward reasoning.
Some organisms have alternative malfunctions ($n_\text{alt} > 0$) that produce
an indistinguishable disease phenotype; these are flagged as ambiguous, and
$n_\text{alt}$ is recorded with every session.

\begin{table}[!ht]
\centering
\caption{Evaluation metrics.}
\label{tab:metrics}
\small
\begin{tabular}{llp{5.8cm}l}
\toprule
Metric & Mode & Definition & Range \\
\midrule
Prediction accuracy & Both & Mean fraction of phenotypes within $\pm 0.10$ of ground truth, averaged over tasks & $[0,1]$ \\
Diagnosis score & Disease & 1.0 correct gene+type; 0.5 correct gene, wrong type; 0.0 wrong gene & $\{0,0.5,1\}$ \\
Rescue fraction & Disease & Fraction of aberrant phenotypes restored to healthy range; a session that never attempts a rescue scores 0 & $[0,1]$ \\
\midrule
\textbf{Composite} & Healthy & Prediction accuracy & $[0,1]$ \\
\textbf{Composite} & Disease & Mean(pred.\ acc., diagnosis, rescue) & $[0,1]$ \\
\bottomrule
\end{tabular}
\end{table}

\paragraph{Agent interface.}
Each model receives the \luca{} API as structured tools via a unified
framework. All providers (Anthropic, OpenAI, Together AI) receive identical
schemas and prompts. A bring-your-own-agent interface supports community
extension. Appendices~\ref{app:prompts} and~\ref{app:session} give the prompts
and a full session as a transcript of tool calls, returned values, and the
budget remaining after each.

\section{Benchmark Validation}
\label{sec:part1}

In this section we test whether \luca{} discriminates between models in the way a useful benchmark should. We ask three questions: 1) does performance vary smoothly with difficulty; 2) does this hold across random organism instances; 3) do model differences reflect genuine reasoning rather than noise?

All organisms are fixed at $N{=}20$ genes.
We run six evaluation conditions: three difficulty tiers (easy, medium, hard) $\times$ two
modes (healthy, disease), with 20 random organisms per condition shared across $10$ models.
This gives $10 \times 6 \times 20 = 1{,}200$ agent--organism sessions in total
(10 models: 7 LLMs and 3 baselines, listed below).\footnote{Two sessions errored at the API layer (gpt-4o seed 15 and gpt-5.1 seed 11, both on medium healthy) and are excluded from aggregate statistics; the corresponding rows use $N=19$ instead of $N=20$.}
We evaluate seven frontier LLMs across three providers:
Claude Opus 4.6 and Claude Sonnet 4.6 (Anthropic),
GPT-5.1, GPT-4o, and GPT-4.1 (OpenAI),
and DeepSeek V3.1 and Qwen 3.5 (open-source, accessed via Together
AI).

In addition, we evaluate three non-LLM baselines.
\textbf{CausalSweep} is a calibration baseline with privileged access to the sigmoid
dynamics model. It BLOCKs and BOOSTs all genes systematically and predicts
outcomes by inverting the sigmoid. LLM agents are not given the underlying
formula.
The \textbf{heuristic} baseline BLOCKs each gene once and predicts by summing individual
effects.
The \textbf{random} baseline predicts $0.5$ for all phenotypes and runs no experiments.
Seed selection and compute details are in Appendix~\ref{app:results}.

Figure~\ref{fig:heatmap} shows composite scores across all six evaluation conditions; standard deviations are in Tables~\ref{tab:app-healthy}--\ref{tab:app-disease} (Appendix~\ref{app:results}).

\begin{figure}[ht]
\centering
\includegraphics[width=0.82\linewidth]{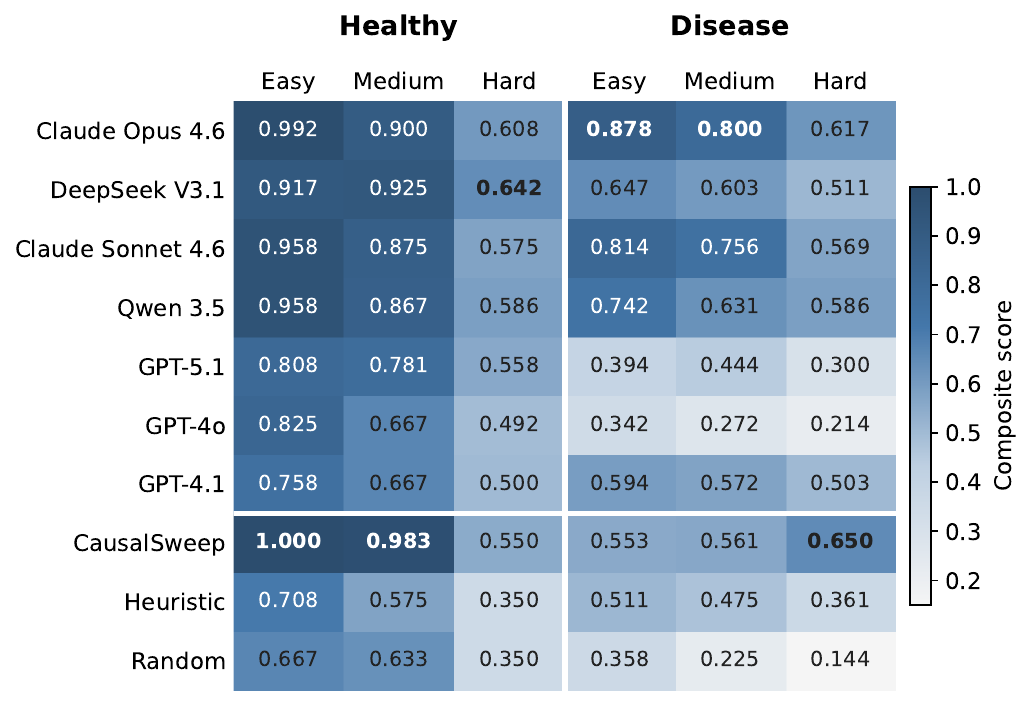}\vspace{-4pt}
\caption{Benchmark validation results.
Composite scores across all six sweeps (easy/medium/hard $\times$ healthy/disease),
$N=20$ seeds per cell, 10 models. Accuracy threshold $\pm0.10$.
Horizontal line separates LLMs from baselines; bold marks the best per column, over all models.
Standard deviations in Tables~\ref{tab:app-healthy}--\ref{tab:app-disease} (Appendix~\ref{app:results}).}
\label{fig:heatmap}
\end{figure}

\paragraph{Performance stratifies by difficulty.}
Medium to hard is the most informative regime: it splits the LLMs into two
groups (Figure~\ref{fig:heatmap}). Opus, DeepSeek, Qwen, and Sonnet cluster
at $0.86$--$0.93$ on medium; GPT-5.1, GPT-4o, and GPT-4.1 sit at
$0.67$--$0.78$. The two prior-free baselines fall below every LLM at both
tiers; CausalSweep, which is given the dynamics model, is the exception and is
treated separately below. On hard tasks, per-seed
variance is high ($\sigma \approx 0.16$--$0.25$) and bimodal: a model tends
to either solve an organism outright or largely fail, rather than degrade
gradually (Appendix~\ref{app:results}). Within each tier, organisms that
are hard for one model are typically hard for the others too, so the
residual variance is structural rather than model-specific.

\paragraph{Limits of brute-force baselines.}
Both heuristic and random converge to a score of $0.35$ on hard healthy organisms,
confirming that additive prediction provides no advantage when gene
interactions are dense.
CausalSweep, which has access to the correct dynamics model, scores $0.983$ on
medium healthy (above all LLMs), confirming that medium prediction is solvable
with correct math alone.
On hard organisms, CausalSweep drops to $0.550$, below the top LLMs.
Its diagnosis score is near chance on easy and medium ($0.00$ and $0.10$),
confirming that diagnosis on these tiers requires reasoning beyond formula
application.

\begin{figure}[ht]
\centering
\includegraphics[width=\linewidth]{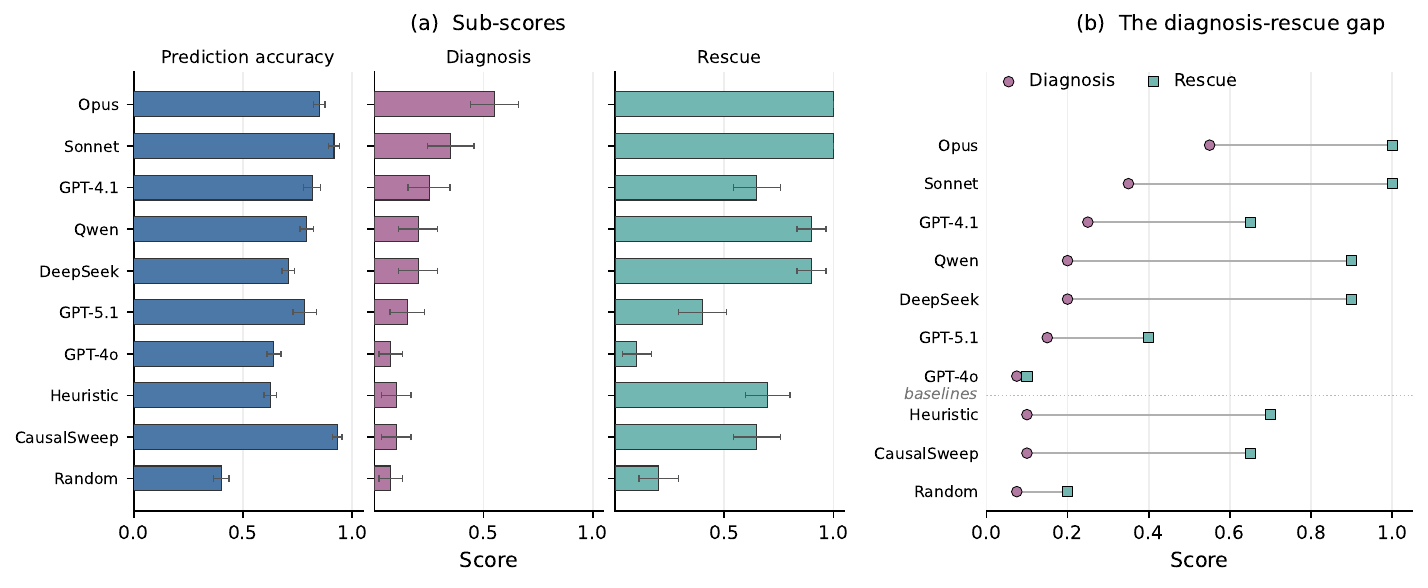}\vspace{-4pt}
\caption{Disease sub-scores on medium organisms (10 models, $N=20$ seeds,
error bars = SE; metrics in Table~\ref{tab:metrics}).
(b) repeats diagnosis and rescue as a per-model gap, ordered by diagnosis,
baselines below the dotted rule: every agent scores higher on rescue than
on diagnosis, including the prior-free baselines.}
\label{fig:subscores}
\end{figure}

\paragraph{Rescue does not imply diagnosis.}
The diagnosis task is where models diverge most.
Prediction accuracy remains relatively high across all models,
but diagnosis and rescue scores diverge dramatically
(Figure~\ref{fig:subscores}).
Claude Opus leads diagnosis on easy and medium tiers, but the diagnosis
ranking does not follow the prediction ranking: the best LLM predictor on
medium disease is Sonnet, the best diagnoser is Opus.
On medium organisms, diagnosis scores span a $7\times$ range across models while prediction scores
span only $1.4\times$, confirming that the two tasks measure distinct capabilities.

This dissociation holds across all model families
(Figure~\ref{fig:subscores}). Every agent scores higher on rescue than on
diagnosis, and Opus, Sonnet, Qwen, and DeepSeek reach $0.90$ or above while diagnosing at most $0.55$. The three baselines reach at most $0.70$, and random only $0.20$. A high rescue score therefore reflects a targeted intervention, not an arbitrary one.
This is structurally possible because phenotypes depend on multiple genes: perturbing
a gene downstream of (or compensating for) the true cause can
restore the affected phenotype without identifying the original malfunction. Because \luca{} borrows the structure of regulatory biology, agents might approach its organisms using priors about how real regulatory networks behave. A structural-prior ablation (Appendix~\ref{app:backwards}) finds little evidence for this: inverting every regulatory sign leaves prediction scores statistically indistinguishable for six of seven models.

\begin{figure}[tbp]
\centering
\includegraphics[width=\linewidth]{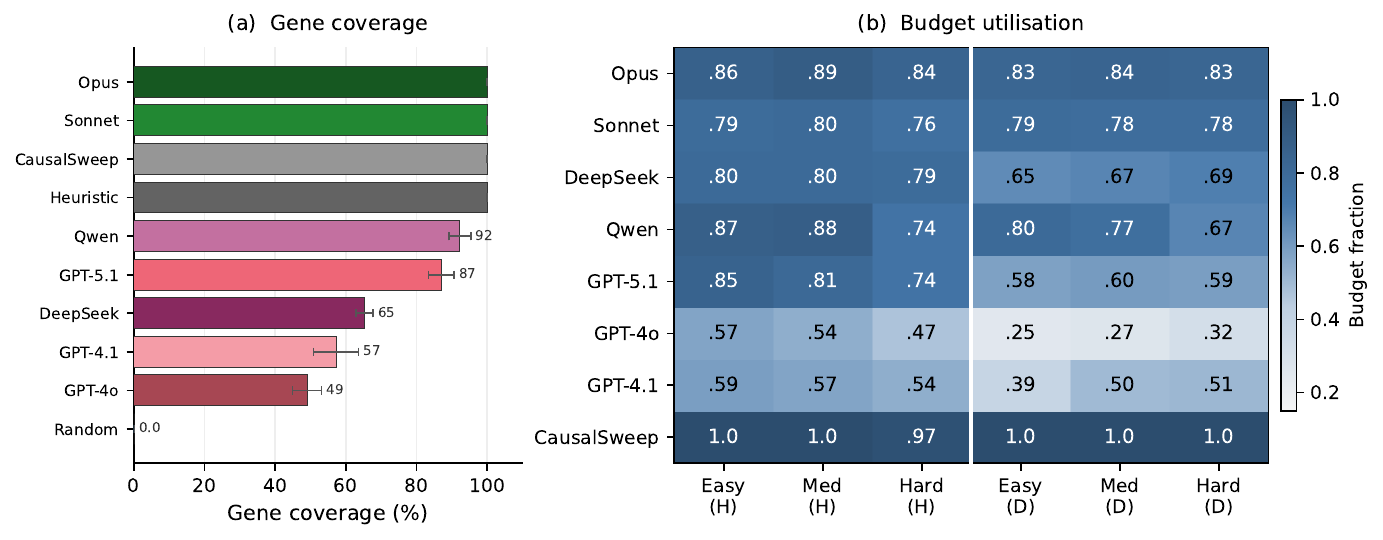}\vspace{-4pt}
\caption{Experimental strategy.
(a) Gene coverage on medium healthy organisms, sorted by fraction of genes tested.
(b) Budget utilisation across all six sweeps (Easy/Medium/Hard $\times$
Healthy/Disease); darker = higher.}
\label{fig:combined}
\end{figure}

\paragraph{Experimental strategy and budget utilisation.}
On medium healthy organisms, models adopt strikingly different experimental strategies
(Figure~\ref{fig:combined}).
Claude models and Qwen run near-exhaustive systematic sweeps with
$92$--$100$\% gene coverage, using $74$--$89$\% of their budget on healthy organisms.
GPT-5.1 achieves comparable coverage ($87$\%) but scores substantially worse,
showing that coverage alone does not translate to accuracy.
DeepSeek uses a similar budget fraction ($80$\%) at lower coverage
($65$\%), indicating less efficient gene selection.
GPT-4o and GPT-4.1 have the lowest LLM coverage ($49$\% and $57$\%
respectively), with high per-seed variance (SE $4$--$7$\%).
In disease mode, GPT-4.1 alone sometimes answers without experimenting at
all: in $3$ of its $60$ disease sessions it ran zero experiments and
submitted a hypothesis directly from the organism description, scoring
zero on diagnosis in each. No other model did this in any session.

\paragraph{Reasoning failures.}
Disease diagnosis also exposed a distinct failure mode
for signal-level malfunctions. A persistently active signal can produce the same perturbation signature as a permanently active gene. Distinguishing the two requires activating the suspected signal: if it is already active, the intervention should have no additional effect. Agents are allowed to perform this test, and signals are explicitly listed as possible malfunction sources.

Across the $56$ LLM sessions on the eight organisms whose true cause is a signal malfunction, agents performed this disambiguating experiment before committing to a diagnosis in $36$ cases. Yet only one session received any diagnosis credit (GPT-4o, 0.5: correct target, wrong malfunction type); all 24 baseline sessions scored zero. Agents therefore often obtained the evidence needed to distinguish the two hypotheses but failed to use it in their diagnosis. 

\section{Scaling Study}
\label{sec:part2}

Section~\ref{sec:part1} fixed organism size at $N{=}20$ and varied
difficulty. We now hold \textbf{medium} difficulty and \textbf{healthy} mode fixed and
ask how performance changes as networks grow, sweeping
$N \in \{20, 40, 80, 160, 320, 640, 1280\}$.

We evaluate four LLMs (Claude Sonnet 4.6, DeepSeek V3.1, Qwen 3.5,
and GPT-4o) as well as CausalSweep, heuristic, and random baselines, across seven
$N$ values (10 seeds each, medium healthy, $\pm 0.10$ threshold).
These four LLMs come from four different model families, so any scaling
claim is not specific to a single training pipeline.
Claude Opus is excluded because at $N \geq 640$ its long internal reasoning
trace fills the output token budget before it can issue a single tool call.
Session budget scales as $B = N + N_{\text{sig}} + 8$, where
$N_{\text{sig}} = \max(3, \lfloor N/7 \rceil)$ is the number of external
signals (3 at $N=20$ to 183 at $N=1280$; see Table~\ref{tab:scaling-rules}).

\begin{figure}[h!]
\centering
\includegraphics[width=\linewidth]{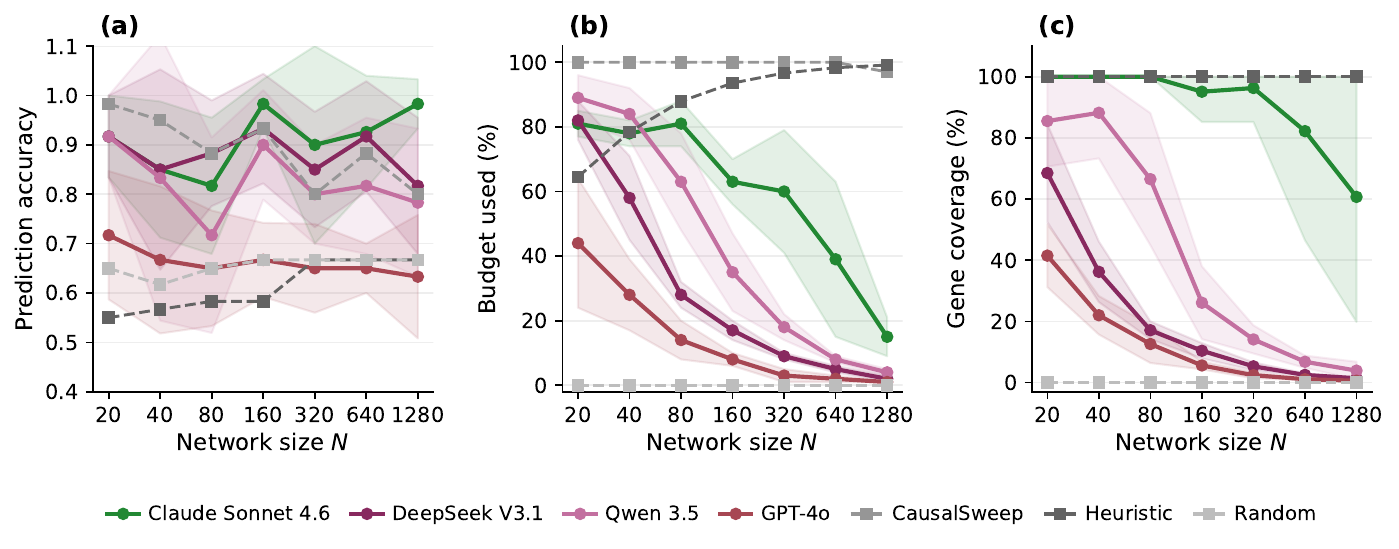}\vspace{-4pt}
\caption{Scaling study (medium healthy, $\pm 0.10$, 10 seeds per cell, 7 models).
Shaded bands = $\pm 1$ std.
(a) Prediction accuracy vs $N$.
(b) Budget utilisation vs $N$.
(c) Gene coverage vs $N$.}
\label{fig:scaling}
\end{figure}

\paragraph{Adaptive reasoning surpasses brute force at scale.}
Figure~\ref{fig:scaling} shows prediction accuracy, budget utilisation, and gene
coverage across all $N$ values (full numbers in
Table~\ref{tab:scaling}, Appendix~\ref{app:results}).
The random baseline converges to the chance-level ceiling for this threshold
($\approx 0.667$\footnote{The $0.667$ floor is structural. Tasks shift one
of three phenotypes by $\geq\!0.10$; the two perturbed genes share a single
phenotype's readout, so their combined effect concentrates on that phenotype
while the other two stay near baseline. A predict-$0.5$ strategy therefore
scores correct on $\approx\!2$ of $3$ phenotypes.}) at $N=160$;
the heuristic baseline reaches it at $N=320$.
Both remain flat through $N=1280$.
The random baseline outperforms the heuristic at $N=20$ ($0.650$ vs $0.550$), confirming
that a baseline built on the wrong assumptions can underperform random guessing.

CausalSweep leads at small $N$ ($0.983$ at $N=20$) but degrades to $0.800$
at $N=1280$, while Sonnet reaches $0.983$. The crossover at $N \approx 160$
is a coverage effect, not a reasoning gap: CausalSweep has the right model
but cannot probe enough of a large network under a fixed budget. Sonnet allocates its budget more selectively.
Sonnet leads from $N \geq 160$, reaching $0.983$ at $N=1280$;
DeepSeek matches Sonnet's accuracy through $N=640$ ($0.917$ vs Sonnet's $0.926$)
before declining to $0.817$ at $N=1280$.
Qwen mirrors DeepSeek at small $N$ but diverges earlier ($0.783$ at $N=1280$).
GPT-4o is the exception: it drops to the baseline floor by $N=40$ and remains
there, offering no advantage over random at any $N \geq 40$.

\begin{figure}[h!]
\centering
\includegraphics[width=0.85\linewidth]{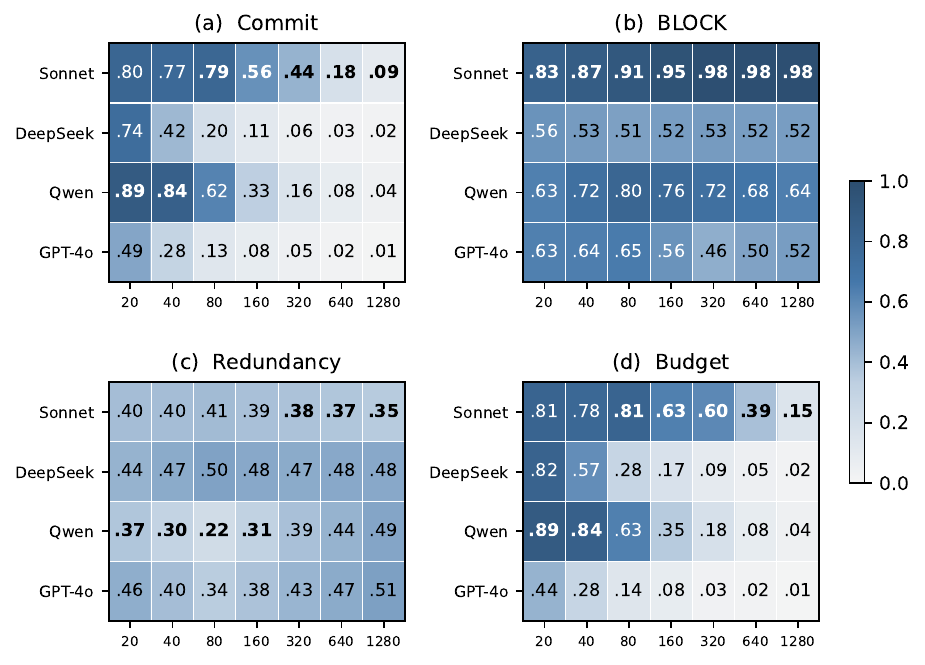}\vspace{-4pt}
\caption{Strategy scaling (four LLMs, all values are fractions, bold = best per column).
(a) Commit: budget fraction before first prediction call.
(b) BLOCK fraction. (c) Redundancy: fraction re-testing a gene.
(d) Budget fraction consumed.}
\label{fig:strategy}
\end{figure}

\paragraph{Strategy explains the divergence.}
The score differences reflect exploration strategies, not implementation
limits. Every model runs its sessions to completion, with the two
exceptions documented in Table~\ref{tab:scaling}
(Figure~\ref{fig:strategy}; Figure~\ref{fig:scaling}, panels~(b)--(c)).
Sonnet adapts its strategy as $N$ grows, running progressively more experiments
($25$ at $N=20$, $\approx\!220$ at $N=1280$), converging to near-exclusive
BLOCK use ($98\%$), and maintaining $61\%$ gene coverage at $N=1280$.
DeepSeek takes the opposite approach: a near-constant $\approx\!30$
experiments regardless of $N$, covering only $\sim$18 genes ($1.4$\% at
$N=1280$), yet scoring $0.917$ at $N=640$.

The most revealing comparison is DeepSeek vs Sonnet.
At $N{=}640$ the two score almost identically ($0.917$ vs $0.926$), but
Sonnet perturbs $526$ of the $640$ genes while DeepSeek perturbs $16$: a
$33\times$ difference in evidence gathered for the same accuracy.
GPT-4o marks the opposite end. It runs the fewest experiments of the four
and its accuracy is indistinguishable from random at every $N \geq 40$.

\paragraph{Implications.}
These results complement Section~\ref{sec:part1} along an orthogonal axis.
Difficulty separates models at fixed $N$; increasing $N$ separates them
by exploration strategy. Together, the two evaluations suggest that closing the
gap requires better experimental design, not just broader exploration.

\section{Conclusion}
\label{sec:conclusion}

We introduced \luca{}, a configurable synthetic world for studying how interactive agents learn unknown systems through experimentation. \luca{} provides controlled access to the full experimental process: which interventions an agent chooses, what it observes, what it infers, and how those inferences translate into prediction, diagnosis, and intervention. Its synthetic construction gives the evaluator complete ground truth while allowing system structure, difficulty, and scale to be varied systematically. Across extensive validation with seven frontier LLMs and three baselines, \luca{} remains discriminative across models and challenging under increasing structural complexity and scale.

Even this initial evaluation reveals behaviours that would be difficult to see from task success alone. Agents frequently rescue malfunctioning systems while diagnosing the wrong fault; models with similar predictive performance adopt strikingly different experimental strategies; and agents sometimes perform the experiment needed to distinguish competing hypotheses but fail to use the resulting evidence. The first echoes the split between predictive success and structural understanding reported for causal-discovery agents~\citep{yang2026causalab}; \luca{} shows that it survives when the agent must also act on the system.

More broadly, we see \luca{} as an instrument for asking questions about experimental agents, rather than a fixed collection of tasks. Its configurable systems and fully observable ground truth make it possible to vary properties of the environment, interventions, and information available to the agent, and then ask how experimental behaviour changes in response. We hope \luca{} provides a useful substrate for developing new agents, evaluation methods, and analyses of experimental reasoning, and invite the community to use and extend it.

\paragraph{Limitations.}
\luca{} studies experimental reasoning in controlled synthetic systems, and several
design choices bound what can be inferred from performance. Organisms use
fixed-point dynamics, excluding oscillatory or multistable behaviour, and the
single-session design does not capture iterative experimentation across
sessions. Invented vocabulary removes potentially useful structural priors,
although our vocabulary ablation finds no measurable performance effect
(Appendix~\ref{app:vocab}).

A further limitation is identifiability. Multiple networks can be consistent
with the same finite set of perturbation observations and disagree on held-out
predictions~\citep{dunn2014defining}. The fixed functional form and experimental
budget partially constrain this ambiguity, but some performance degradation may
reflect underdetermination rather than agent behaviour alone.

\paragraph{Future work.}
These limitations also suggest natural extensions to this synthetic environment. The
configurable generator enables systematic study of alternative dynamics,
structural priors, and identifiability, while multi-session protocols could
extend \luca{} to iterative experimentation. The controlled setting also enables
finer-grained trajectory analysis: which experiments are informative, whether
agents select them, and how subsequent predictions and diagnoses change after
informative evidence is observed.

\section*{Broader Impact}
\luca{} shows that successful intervention can coexist with incorrect diagnosis.
Although this result is established only in synthetic systems, it illustrates
a limitation of outcome-only evaluation: successful action alone does not
establish how an agent gathered, interpreted, or used evidence. This distinction
is particularly relevant when evaluating agents intended to support
experimental science.

\section*{Data and Code Availability}
Benchmark code: \url{https://github.com/RelationRx/luca} (CC BY-NC 4.0).

\bibliographystyle{unsrtnat}
\bibliography{references}

\appendix
\section*{Appendix}
\addcontentsline{toc}{section}{Appendix}

\section{Organism Generation Details}
\label{app:generation}

Each organism is generated deterministically from a \texttt{(seed, tier, $N$)} triple.
This appendix reports the resulting per-tier statistics
(\S\ref{app:gen-stats}), the generation pipeline that produces them
(\S\ref{app:pipeline}), and the scaling rules that extend the pipeline
to larger $N$ (\S\ref{app:scaling}).

\subsection{Per-tier organism statistics}
\label{app:gen-stats}

Difficulty tiers vary along three axes: observability (the fraction of
genes whose states drive phenotype readouts), GRN density, and the
presence of latent genes (genes that participate in the dynamics but
cannot be perturbed by the agent). Table~\ref{tab:stats} reports the
resulting structural statistics over $20$ pre-screened seeds per tier.
Two patterns to note: no prediction task at any tier is solvable by
linear superposition within the $\pm 0.10$ scoring threshold, and the
additive residual grows with tier, so the prediction tasks are genuinely
non-additive; the count of alternative malfunctions $n_\text{alt}$ also
rises with tier.

\begin{table}[h]
\centering
\caption{Per-tier organism statistics (20 pre-screened seeds per tier, $N=20$ genes, $M=4$ modules, 3 phenotypes, 3 signals).
Agent-targetable = non-latent genes the agent can perturb.
Readout genes = subset with direct phenotype readout connections.
Additive task rate: fraction of prediction tasks solvable by linear superposition
within the $\pm 0.10$ scoring threshold.
Additive residual: mean over seeds of the largest such error per organism
($\pm$ population std).
$n_\text{alt}$: number of alternative malfunctions producing an observationally
equivalent disease phenotype (mean / max over seeds).}
\label{tab:stats}
\small
\begin{tabular}{lccc}
\toprule
 & Easy & Medium & Hard \\
\midrule
Agent-targetable genes    & 20        & 20        & 15        \\
Readout genes             & 16 (80\%) & 12 (60\%) & 10 (50\%) \\
Latent genes              & 0         & 0         & 5 (25\%)  \\
GRN edges (range / mean)  & 26--42 / 34 & 58--75 / 67 & 87--121 / 102 \\
Additive task rate (mean) & 0.00      & 0.00      & 0.00      \\
Additive residual (mean)  & $0.166 \pm 0.033$ & $0.225 \pm 0.056$ & $0.309 \pm 0.082$ \\
$n_\text{alt}$ (mean / max) & 0.05 / 1 & 0.50 / 2 & 1.25 / 4 \\
Seeds passing verification & 20/20    & 20/20     & 20/20     \\
\bottomrule
\end{tabular}
\end{table}

\subsection{Generation pipeline}
\label{app:pipeline}

The 13-step pipeline (Table~\ref{tab:pipeline}) constructs each
organism from its seed. Steps 1--9 build the network topology; steps
11--13 enforce four functional guarantees that prevent degenerate
organisms: no silent observable genes, non-additive prediction tasks,
relevant latent genes, and calibrated phenotype baselines.

\begin{table}[h]
\centering
\caption{Organism generation pipeline. Steps 11--13 enforce the four
non-degeneracy guarantees: no silent observable genes, non-additive prediction
tasks, relevant latent genes, and calibrated phenotype baselines.}
\label{tab:pipeline}
\resizebox{\textwidth}{!}{%
\begin{tabular}{clp{6.5cm}}
\toprule
Step & Action & Key guarantee \\
\midrule
1  & Assign gene names (G001--GXXX) & \\
2  & Partition genes into $M$ modules & $\geq 1$ gene per module \\
3  & Sample GRN edges (Erd\H{o}s--R\'{e}nyi) & Density independent of $N$ \\
4  & Inject feed-forward loops & Count $\propto N$ \\
5  & Inject feedback loops & Count $\propto N$ \\
6  & Assign signals and target genes & Coverage $\propto N$ \\
7  & Sample per-gene biases & \\
8  & Select observable genes & $\geq 1$ per module \\
9  & Select latent gene candidates & From non-observable pool \\
10 & Build sigmoid dynamics; simulate steady state & \\
11 & Build phenotype readouts & Every observable gene in one readout \\
12 & Latent gene relevance check & Each latent gene shifts $\geq 1$ phenotype \\
13 & Re-centre phenotype biases & Every phenotype baseline $= 0.5$ \\
\bottomrule
\end{tabular}%
}
\end{table}

\subsection{Scaling rules}
\label{app:scaling}

All structural parameters scale automatically with $N$ via the formulas
in Table~\ref{tab:scaling-rules}; the agent interface is identical at
every scale, so the same evaluation harness runs from $N=20$ to
$N=1{,}280$.

\begin{table}[h]
\centering
\caption{Scaling rules for organism parameters.}
\label{tab:scaling-rules}
\small
\begin{tabular}{lccccc}
\toprule
Parameter & Formula & $N=20$ & $N=40$ & $N=80$ & $N=160$ \\
\midrule
Modules $M$      & $\max(4,\lfloor N/5 \rceil)$    & 4  & 8  & 16 & 32 \\
Signals          & $\max(3,\lfloor N/7 \rceil)$    & 3  & 6  & 11 & 23 \\
Signal max targets & $\max(3,\lfloor N/7 \rceil)$ & 3  & 6  & 11 & 23 \\
Phenotypes       & fixed                            & 3  & 3  & 3  & 3  \\
Prediction tasks & fixed                            & 2  & 2  & 2  & 2  \\
Budget $B$       & $N + N_\text{sig} + 8$           & 31 & 54 & 99 & 191 \\
\bottomrule
\end{tabular}
\end{table}

\FloatBarrier
\clearpage

\section{Agent Prompts}
\label{app:prompts}

System prompts are constructed dynamically by \texttt{\_build\_system\_prompt(disease\_mode, active\_tasks)}.
Both modes share a \textsc{critical} block asserting that all vocabulary is invented and prior
biological knowledge is irrelevant.
The two canonical prompts are reproduced below.

\subsection{Healthy mode (predict only)}

\begin{quote}
\small\ttfamily
You are a computational biologist studying a \textit{de novo} synthetic organism.

CRITICAL: All gene names (e.g.\ G001, G042), signal labels (e.g.\ ``nutrient-X''), and
phenotype names (e.g.\ ``growth-rate'') are INVENTED VOCABULARY. They do NOT correspond
to any real biological system. Prior biological knowledge is IRRELEVANT and may mislead
you. The only valid path to understanding this organism is systematic causal reasoning
from perturbation experiments.

PRIMARY GOALS (in order of importance):\\
1. Phenotype prediction --- accurately predict phenotype outcomes under novel multi-gene
interventions. This requires building a real causal model, not guessing.\\
2. Causal hypothesis --- submit a concise, evidence-based hypothesis naming which genes
regulate which phenotypes and how.

WORKFLOW:\\
1. Call get\_organism\_info() to understand the gene set, signals, phenotypes, and baseline.\\
2. Run perturbation experiments (BLOCK, BOOST, DAMPEN, signal activations) to map
gene$\to$phenotype causal relationships.\\
3. When you have a working causal model, call get\_next\_prediction\_task() for one task
at a time; submit via submit\_prediction().\\
4. Submit your final causal hypothesis via submit\_hypothesis().\\
5. Submit your reconstructed network via submit\_network().

Strategy tips:\\
- Single-gene BLOCK/BOOST experiments first.\\
- BUDGET MANAGEMENT IS CRITICAL: Reserve $\geq$3 budget units for prediction tasks.\\
- Prediction tasks involve two simultaneous cross-module interventions --- reason from
your causal model.\\
- Do not guess 0.5 --- every prediction should be grounded in experimental evidence.
\end{quote}

\subsection{Disease mode (diagnose + rescue + predict)}

\begin{quote}
\small\ttfamily
You are a computational biologist studying a \textit{de novo} synthetic organism that is
exhibiting a diseased state.

CRITICAL: [as above]

DISEASE CONTEXT: This cell exhibits aberrant phenotypes compared to its healthy baseline
ranges. The regulatory network has a single malfunction --- either a node locked high
(LOCK\_HI), a node locked low (LOCK\_LO), or a persistently active external signal
(PERSIST). Your job is to identify it and restore healthy phenotypes.

PRIMARY GOALS (in order of importance):\\
1. Malfunction diagnosis --- identify which gene or signal is broken and how
(LOCK\_HI / LOCK\_LO / PERSIST). Submit via submit\_malfunction\_hypothesis().\\
2. Rescue --- design an intervention that restores aberrant phenotypes. Test via test\_rescue().\\
3. Phenotype prediction --- as above.\\
4. Causal hypothesis --- as above.

WORKFLOW:\\
1. Call get\_organism\_info().\\
2. Run perturbation experiments targeting genes affecting aberrant phenotypes.\\
3. Submit malfunction hypothesis via submit\_malfunction\_hypothesis() (one-time, irreversible).\\
4. Test rescue interventions via test\_rescue().\\
5--7. Prediction, hypothesis, network as healthy mode.

Strategy tips:\\
- LOCK\_HI: BLOCK/DAMPEN should reduce overactive phenotypes.\\
- LOCK\_LO: BOOST should restore under-active phenotypes.\\
- PERSIST: signal always on --- BLOCK direct targets.\\
- BUDGET MANAGEMENT IS CRITICAL: Reserve $\geq$4 units for rescue + prediction.\\
- Commit to your hypothesis only when evidence is clear --- you cannot revise it.
\end{quote}

\section{A Worked Session}
\label{app:session}

The interface described in Section~\ref{sec:interface} is easiest to read as a
transcript. This appendix reproduces one complete disease-mode session:
Claude Opus 4.6 on an easy-tier organism, seed 0, budget $B{=}31$, of which the
agent used 30 units. Tool calls and returned values are shown as they occurred.
Phenotype values are rounded to three decimals, the 20-element gene list is
abridged, two descriptive fields are omitted from the first response, and long
values are wrapped to fit the page; nothing else is edited.

The session opens with a free call that costs no budget. It reports the healthy
reference ranges and names which phenotype is currently aberrant.

{\small
\begin{verbatim}
>>> get_organism_info()
{"organism_id": "org_00000000",
 "gene_names": ["G001", "G002", ..., "G020"],
 "signals": [{"label": "cascade-B", "mode": "activates"},
             {"label": "flux-theta", "mode": "activates"},
             {"label": "flux-epsilon", "mode": "inhibits"}],
 "observable_phenotypes": ["energy-state", "displacement-score",
                           "standby-level"],
 "baseline_phenotypes": {"energy-state": 0.504,
                         "displacement-score": 0.658,
                         "standby-level": 0.502},
 "healthy_phenotype_ranges": {"energy-state": [0.35, 0.65],
                              "displacement-score": [0.35, 0.65],
                              "standby-level": [0.35, 0.65]},
 "aberrant_phenotypes": ["displacement-score"],
 "budget_remaining": 31, "budget_total": 31}
\end{verbatim}
}

Each experiment costs one unit. The agent first blocks genes in groups of five
to localise the fault. Every return value carries the remaining budget, so the
cost of a strategy is visible to the agent as it spends.

{\small
\begin{verbatim}
>>> run_experiment({"G001": {"type": "BLOCK"}, "G002": {"type": "BLOCK"},
                    "G003": {"type": "BLOCK"}, "G004": {"type": "BLOCK"},
                    "G005": {"type": "BLOCK"}})
{"phenotypes": {"energy-state": 0.344, "displacement-score": 0.663,
                "standby-level": 0.205},
 "status": "max_steps_reached", "steps": 500, "budget_remaining": 30}
\end{verbatim}
}

Three further group experiments cover the remaining genes. Two of the four
groups shift the aberrant phenotype, and the agent switches to single genes.

{\small
\begin{verbatim}
>>> run_experiment({"G006": {"type": "BLOCK"}})
{"phenotypes": {"energy-state": 0.486, "displacement-score": 0.562,
                "standby-level": 0.513},
 "status": "max_steps_reached", "steps": 500, "budget_remaining": 26}
\end{verbatim}
}

After 21 experiments the agent commits to a diagnosis. The commitment is
one-way: it unlocks \texttt{test\_rescue()} and cannot be revised, and the score
is withheld until the session ends.

{\small
\begin{verbatim}
>>> submit_malfunction_hypothesis(target_name="G012",
                                  malfunction_type="LOCK_HI")
{"stored": true,
 "note": "Hypothesis locked. test_rescue() is now available.
          Score revealed at session end."}
\end{verbatim}
}

The agent then proposes a corrective intervention, which restores every
phenotype to its healthy range.

{\small
\begin{verbatim}
>>> test_rescue({"G012": {"type": "BLOCK"}})
{"interventions": {"G012": {"type": "BLOCK"}},
 "phenotypes": {"energy-state": 0.501, "displacement-score": 0.470,
                "standby-level": 0.515},
 "phenotype_in_healthy_range": {"energy-state": true,
                                "displacement-score": true,
                                "standby-level": true},
 "restored_fraction": 1.0, "mse_vs_healthy": 0.0004,
 "budget_remaining": 9}
\end{verbatim}
}

At session end the scorer reveals the ground truth. The malfunction was G018
locked low, not G012 locked high, so the diagnosis scores $0.0$ with the
feedback \texttt{wrong target}, while the rescue scores $1.0$. Blocking G012
compensated for a fault the agent had not identified. This session is a single
instance of the dissociation reported in Section~\ref{sec:part1}: the agent
acted correctly on the system while holding an incorrect model of it. An
outcome-only score would record it as a success.

\section{Extended Results}
\label{app:results}

This appendix provides full numerical results for the benchmark validation
(Section~\ref{sec:part1}) and scaling study (Section~\ref{sec:part2}),
including per-tier breakdowns, strategy metrics, and network reconstruction analysis.

\begin{table}[h]
\centering
\caption{Healthy mode composite scores (mean $\pm$ std, $N=20$ seeds;
std is population standard deviation $\sigma_n$).
Accuracy threshold $\pm 0.10$. Bold: best per column, over all models.
$\dagger$: non-LLM baselines.}
\label{tab:app-healthy}
\small
\begin{tabular}{lccc}
\toprule
Model & Easy & Medium & Hard \\
\midrule
claude-opus-4-6   & 0.992 $\pm$ 0.036 & 0.900 $\pm$ 0.111 & 0.608 $\pm$ 0.192 \\
deepseek-v3.1     & 0.917 $\pm$ 0.124 & 0.925 $\pm$ 0.098 & \textbf{0.642} $\pm$ 0.161 \\
claude-sonnet-4-6 & 0.958 $\pm$ 0.089 & 0.875 $\pm$ 0.128 & 0.575 $\pm$ 0.250 \\
qwen-3.5          & 0.958 $\pm$ 0.072 & 0.867 $\pm$ 0.113 & 0.586 $\pm$ 0.178 \\
gpt-5.1           & 0.808 $\pm$ 0.142 & 0.781 $\pm$ 0.133 & 0.558 $\pm$ 0.206 \\
gpt-4o            & 0.825 $\pm$ 0.123 & 0.667 $\pm$ 0.132 & 0.492 $\pm$ 0.201 \\
gpt-4.1           & 0.758 $\pm$ 0.134 & 0.667 $\pm$ 0.075 & 0.500 $\pm$ 0.253 \\
\midrule
causal-sweep$^\dagger$ & \textbf{1.000} $\pm$ 0.000 & \textbf{0.983} $\pm$ 0.050 & 0.550 $\pm$ 0.205 \\
heuristic$^\dagger$ & 0.708 $\pm$ 0.089 & 0.575 $\pm$ 0.179 & 0.350 $\pm$ 0.196 \\
random$^\dagger$    & 0.667 $\pm$ 0.000 & 0.633 $\pm$ 0.085 & 0.350 $\pm$ 0.223 \\
\bottomrule
\end{tabular}
\end{table}

\begin{table}[h]
\centering
\caption{Disease mode composite scores (mean $\pm$ std, $N=20$ seeds;
std is population standard deviation $\sigma_n$).
Composite = mean(pred\_acc, diagnosis, rescue). Accuracy threshold $\pm 0.10$.
Bold: best per column, over all models.
Unattempted tasks scored as 0. $\dagger$: non-LLM baselines.}
\label{tab:app-disease}
\small
\begin{tabular}{lccc}
\toprule
Model & Easy & Medium & Hard \\
\midrule
claude-opus-4-6        & \textbf{0.878} $\pm$ 0.165 & \textbf{0.800} $\pm$ 0.159 & 0.617 $\pm$ 0.191 \\
claude-sonnet-4-6      & 0.814 $\pm$ 0.184 & 0.756 $\pm$ 0.160 & 0.569 $\pm$ 0.244 \\
qwen-3.5               & 0.742 $\pm$ 0.147 & 0.631 $\pm$ 0.197 & 0.586 $\pm$ 0.220 \\
deepseek-v3.1          & 0.647 $\pm$ 0.113 & 0.603 $\pm$ 0.171 & 0.511 $\pm$ 0.227 \\
gpt-4.1                & 0.594 $\pm$ 0.283 & 0.572 $\pm$ 0.262 & 0.503 $\pm$ 0.264 \\
gpt-5.1                & 0.394 $\pm$ 0.199 & 0.444 $\pm$ 0.242 & 0.300 $\pm$ 0.215 \\
gpt-4o                 & 0.342 $\pm$ 0.164 & 0.272 $\pm$ 0.127 & 0.214 $\pm$ 0.134 \\
\midrule
causal-sweep$^\dagger$ & 0.553 $\pm$ 0.147 & 0.561 $\pm$ 0.201 & \textbf{0.650} $\pm$ 0.254 \\
heuristic$^\dagger$    & 0.511 $\pm$ 0.290 & 0.475 $\pm$ 0.204 & 0.361 $\pm$ 0.277 \\
random$^\dagger$       & 0.358 $\pm$ 0.188 & 0.225 $\pm$ 0.173 & 0.144 $\pm$ 0.126 \\
\bottomrule
\end{tabular}
\end{table}

\begin{table}[h]
\centering
\caption{Disease mode sub-score breakdown by difficulty tier ($N=20$ seeds each).
Composite = mean(pred\_acc, diagnosis, rescue). Accuracy threshold $\pm 0.10$.
Bold: best per column, over all models, within each tier.
Unattempted tasks scored as 0. $\dagger$: non-LLM baselines.}
\label{tab:app-disease-sub}
\small
\begin{tabular}{lcccc}
\toprule
Model & Pred acc & Diagnosis & Rescue & Composite \\
\midrule
\multicolumn{5}{l}{\textit{Easy disease}} \\
\midrule
claude-opus-4-6        & 0.933 & \textbf{0.700} & \textbf{1.000} & \textbf{0.878} \\
claude-sonnet-4-6      & 0.892 & 0.550 & \textbf{1.000} & 0.814 \\
qwen-3.5               & 0.925 & 0.300 & \textbf{1.000} & 0.742 \\
deepseek-v3.1          & 0.792 & 0.200 & 0.950 & 0.647 \\
gpt-4.1                & 0.833 & 0.350 & 0.600 & 0.594 \\
causal-sweep$^\dagger$ & \textbf{0.958} & 0.000 & 0.700 & 0.553 \\
heuristic$^\dagger$    & 0.708 & 0.300 & 0.525 & 0.511 \\
gpt-5.1                & 0.833 & 0.100 & 0.250 & 0.394 \\
random$^\dagger$       & 0.525 & 0.025 & 0.525 & 0.358 \\
gpt-4o                 & 0.700 & 0.075 & 0.250 & 0.342 \\
\midrule
\multicolumn{5}{l}{\textit{Medium disease}} \\
\midrule
claude-opus-4-6        & 0.850 & \textbf{0.550} & \textbf{1.000} & \textbf{0.800} \\
claude-sonnet-4-6      & 0.917 & 0.350 & \textbf{1.000} & 0.756 \\
qwen-3.5               & 0.792 & 0.200 & 0.900 & 0.631 \\
deepseek-v3.1          & 0.708 & 0.200 & 0.900 & 0.603 \\
gpt-4.1                & 0.817 & 0.250 & 0.650 & 0.572 \\
causal-sweep$^\dagger$ & \textbf{0.933} & 0.100 & 0.650 & 0.561 \\
heuristic$^\dagger$    & 0.625 & 0.100 & 0.700 & 0.475 \\
gpt-5.1                & 0.783 & 0.150 & 0.400 & 0.444 \\
gpt-4o                 & 0.642 & 0.075 & 0.100 & 0.272 \\
random$^\dagger$       & 0.400 & 0.075 & 0.200 & 0.225 \\
\midrule
\multicolumn{5}{l}{\textit{Hard disease}} \\
\midrule
causal-sweep$^\dagger$ & \textbf{0.692} & \textbf{0.550} & 0.708 & \textbf{0.650} \\
claude-opus-4-6        & 0.558 & 0.350 & \textbf{0.942} & 0.617 \\
qwen-3.5               & 0.517 & 0.400 & 0.842 & 0.586 \\
claude-sonnet-4-6      & 0.592 & 0.300 & 0.817 & 0.569 \\
deepseek-v3.1          & 0.417 & 0.300 & 0.817 & 0.511 \\
gpt-4.1                & 0.658 & 0.275 & 0.575 & 0.503 \\
heuristic$^\dagger$    & 0.458 & 0.150 & 0.475 & 0.361 \\
gpt-5.1                & 0.517 & 0.225 & 0.158 & 0.300 \\
gpt-4o                 & 0.433 & 0.100 & 0.108 & 0.214 \\
random$^\dagger$       & 0.300 & 0.075 & 0.058 & 0.144 \\
\bottomrule
\end{tabular}
\end{table}

Table~\ref{tab:app-disease-sub} decomposes the disease-mode composite into prediction accuracy, diagnosis, and rescue for each tier; it is the source for the rescue--diagnosis gap discussed in Section~\ref{sec:part1}.

\paragraph{Correction to the rescue metric.}
An earlier release of the scorer did not compute the rescue fraction as
Table~\ref{tab:metrics} defines it. It divided the phenotypes inside their
healthy range by \emph{all} phenotypes rather than by the aberrant ones, so on
an organism with one aberrant phenotype of three an intervention that changed
nothing scored $2/3$. Every rescue score in this paper uses the corrected
definition, as does the current release. Prediction and diagnosis are
unaffected, and so is the ordering the rescue--diagnosis gap rests on: the
correction lowers rescue most for the weakest agents, whose apparent success
had been recording inaction rather than repair.

Table~\ref{tab:app-budget} reports budget utilisation across all six Part~1 sweeps. It lists the seven LLMs only: the heuristic baseline is deterministic and random runs no experiments, so neither varies. Figure~\ref{fig:combined}(b) additionally shows CausalSweep, which does vary.

\begin{table}[h]
\centering
\caption{Mean budget utilisation fraction (experiments run / total budget) across all six sweeps (7 LLMs).}
\label{tab:app-budget}
\small
\begin{tabular}{lcccccc}
\toprule
Model & Easy (H) & Med (H) & Hard (H) & Easy (D) & Med (D) & Hard (D) \\
\midrule
claude-opus-4-6   & 0.86 & 0.89 & 0.84 & 0.83 & 0.84 & 0.83 \\
claude-sonnet-4-6 & 0.79 & 0.80 & 0.76 & 0.79 & 0.78 & 0.78 \\
deepseek-v3.1     & 0.80 & 0.80 & 0.79 & 0.65 & 0.67 & 0.69 \\
qwen-3.5          & 0.87 & 0.88 & 0.74 & 0.80 & 0.77 & 0.67 \\
gpt-5.1           & 0.85 & 0.81 & 0.74 & 0.58 & 0.60 & 0.59 \\
gpt-4o            & 0.57 & 0.54 & 0.47 & 0.25 & 0.27 & 0.32 \\
gpt-4.1           & 0.59 & 0.57 & 0.54 & 0.39 & 0.50 & 0.51 \\
\bottomrule
\end{tabular}
\end{table}

Table~\ref{tab:app-strategy} breaks down experimental strategy on medium healthy organisms.

\begin{table}[h]
\centering
\caption{Experimental strategy analysis on medium healthy organisms
(7 LLMs and 2 baselines, $N=20$ seeds).
SingleGene\%: fraction of experiments targeting a single gene.
Coverage\%: fraction of observable genes tested at least once.
$\dagger$: non-LLM baselines; random runs no experiments and is omitted.}
\label{tab:app-strategy}
\small
\begin{tabular}{lccc}
\toprule
Model & Budget frac & SingleGene\% & Coverage\% \\
\midrule
causal-sweep$^\dagger$ & 1.00 & 100.0\% & 100.0\% \\
claude-opus-4-6   & 0.89 & 100.0\% & 100.0\% \\
claude-sonnet-4-6 & 0.80 & 76.3\% & 100.0\% \\
qwen-3.5          & 0.88 & 100.0\% &  92.2\% \\
gpt-5.1           & 0.81 & 95.8\% &  87.0\% \\
deepseek-v3.1     & 0.80 & 99.6\% &  65.2\% \\
heuristic$^\dagger$ & 0.65 & 100.0\% & 100.0\% \\
gpt-4.1           & 0.57 & 89.1\% &  57.2\% \\
gpt-4o            & 0.54 & 95.0\% &  49.0\% \\
\bottomrule
\end{tabular}
\end{table}

\subsection{Scaling Study Results}
\label{app:scaling-results}

\begin{table}[h]
\centering
\caption{Scaling study prediction accuracy (mean $\pm$ std, medium healthy,
$\pm 0.10$ threshold, 10 seeds per cell unless noted;
std is population standard deviation $\sigma_n$).
Budget $B = N + N_{\text{sig}} + 8$.
Bold: best per row, over all models.
$\dagger$: heuristic $N=1280$ stopped at 5 seeds (structural floor confirmed).
Sonnet $N=640$ completed on 9 of 10 seeds (one API error).}
\label{tab:scaling}
\resizebox{\textwidth}{!}{%
\begin{tabular}{rrlllllll}
\toprule
$N$ & Budget & CausalSweep & Sonnet & DeepSeek & Qwen & GPT-4o & Heuristic & Random \\
\midrule
20   & 31   & $\mathbf{0.983} \pm 0.050$ & $0.917 \pm 0.083$ & $0.917 \pm 0.083$ & $0.917 \pm 0.083$ & $0.717 \pm 0.130$ & $0.550 \pm 0.150$ & $0.650 \pm 0.050$ \\
40   & 54   & $\mathbf{0.950} \pm 0.076$ & $0.850 \pm 0.138$ & $0.850 \pm 0.203$ & $0.833 \pm 0.289$ & $0.667 \pm 0.149$ & $0.567 \pm 0.082$ & $0.617 \pm 0.107$ \\
80   & 99   & $0.883 \pm 0.150$ & $0.817 \pm 0.138$ & $\mathbf{0.883} \pm 0.107$ & $0.717 \pm 0.198$ & $0.650 \pm 0.117$ & $0.583 \pm 0.201$ & $0.650 \pm 0.050$ \\
160  & 191  & $0.933 \pm 0.082$ & $\mathbf{0.983} \pm 0.050$ & $0.933 \pm 0.111$ & $0.900 \pm 0.111$ & $0.667 \pm 0.075$ & $0.583 \pm 0.154$ & $0.667 \pm 0.000$ \\
320  & 374  & $0.800 \pm 0.145$ & $\mathbf{0.900} \pm 0.200$ & $0.850 \pm 0.117$ & $0.800 \pm 0.100$ & $0.650 \pm 0.090$ & $0.667 \pm 0.129$ & $0.667 \pm 0.000$ \\
640  & 739  & $0.883 \pm 0.107$ & $\mathbf{0.926} \pm 0.114$ & $0.917 \pm 0.112$ & $0.817 \pm 0.138$ & $0.650 \pm 0.050$ & $0.667 \pm 0.075$ & $0.667 \pm 0.000$ \\
1280 & 1471 & $0.800 \pm 0.125$ & $\mathbf{0.983} \pm 0.050$ & $0.817 \pm 0.138$ & $0.783 \pm 0.150$ & $0.633 \pm 0.125$ & $0.667 \pm 0.000^{\dagger}$ & $0.667 \pm 0.000$ \\
\bottomrule
\end{tabular}%
}
\end{table}

\subsection{Network Reconstruction}
\label{app:network}

We score each model's inferred gene$\to$phenotype edge set against ground truth
(Table~\ref{tab:app-network}).
Edges are \texttt{(gene, phenotype, direction)} triples inferred from phenotype
shifts $>0.05$ on single-gene BLOCK/BOOST experiments.

Claude Opus achieves high recall by over-inferring edges, trading precision for
coverage: on hard organisms it infers 34.0 edges on average against a ground
truth of $\sim$12, while precision collapses to 0.27.
The correlation between per-seed F1 and prediction accuracy is weak and
inconsistent across models ($r$ ranges from $-0.43$ to $+0.46$), and is
\emph{negative} for Opus on medium, the model with the highest F1 here and the
second-highest prediction accuracy of any LLM at that tier
(Table~\ref{tab:app-healthy}).
This suggests prediction performance is not driven by explicit edge
reconstruction and that F1 is a poor proxy for task-relevant causal understanding.

\begin{table}[h]
\centering
\caption{Network reconstruction metrics (mean over $N=20$ seeds, medium and hard
healthy).
Edges = mean inferred \texttt{(gene, phenotype, direction)} triples per session;
ground truth $\approx 12$ per organism at both tiers.
$r$ = Pearson correlation between per-seed F1 and prediction accuracy.}
\label{tab:app-network}
\resizebox{\textwidth}{!}{%
\begin{tabular}{lcccccccccc}
\toprule
& \multicolumn{5}{c}{Medium} & \multicolumn{5}{c}{Hard} \\
\cmidrule(lr){2-6} \cmidrule(lr){7-11}
Model & Prec & Rec & F1 & Edges & $r$ & Prec & Rec & F1 & Edges & $r$ \\
\midrule
claude-opus-4-6   & 0.640 & 0.933 & 0.754 & 18.1 & $-0.30$ & 0.265 & 0.752 & 0.390 & 34.0 & $+0.18$ \\
claude-sonnet-4-6 & 0.675 & 0.775 & 0.710 & 14.1 & $+0.09$ & 0.277 & 0.675 & 0.390 & 30.0 & $+0.26$ \\
qwen-3.5          & 0.615 & 0.742 & 0.660 & 14.8 & $+0.22$ & 0.351 & 0.740 & 0.464 & 23.8 & $+0.29$ \\
deepseek-v3.1     & 0.583 & 0.662 & 0.611 & 15.0 & $+0.16$ & 0.290 & 0.555 & 0.372 & 21.2 & $-0.43$ \\
gpt-5.1           & 0.570 & 0.754 & 0.644 & 15.3 & $+0.01$ & 0.257 & 0.719 & 0.376 & 30.7 & $-0.02$ \\
gpt-4o            & 0.325 & 0.246 & 0.270 &  5.8 & $-0.16$ & 0.128 & 0.205 & 0.153 & 10.1 & $+0.02$ \\
gpt-4.1           & 0.153 & 0.112 & 0.109 &  3.0 & $+0.21$ & 0.227 & 0.441 & 0.294 & 20.1 & $+0.46$ \\
\bottomrule
\end{tabular}%
}
\end{table}

Per-seed score distributions are available in the project repository.

\FloatBarrier
\clearpage

\section{Robustness Ablations}
\label{app:robustness}

Two ablations probe the benchmark's design. The first
(\S\ref{app:backwards}) tests whether prediction performance reflects
observation-driven inference or transferred priors from real-biology GRN
literature. The second (\S\ref{app:vocab}) tests whether scores are
confounded by gene-name vocabulary or by the system prompt's stance on
biological knowledge.

\subsection{Structural-prior ablation (backwards organisms)}
\label{app:backwards}

\paragraph{Design.}
For every medium-tier seed we constructed a \emph{backwards} variant by
inverting every signed quantity: GRN edge weights $w_{ij}$, gene
biases $b_i$, signal modes, and phenotype readout weights. Topology,
magnitudes, and noise were preserved; phenotype baselines were
re-centred so $\sigma(0) \approx 0.5$ at steady state. The backwards
organism violates real-biology priors (e.g.\ ``blocking an activator
raises its downstream targets'') while keeping the same observable
input--output structure.

Twenty canonical medium-tier seeds were evaluated in both directions in
healthy mode. Analysis is paired per seed:
$\Delta = \text{forward} - \text{backward}$. We report paired Wilcoxon
signed-rank tests and percentile-bootstrap $95\%$ CIs. Seven LLMs
(Opus~4.6, Sonnet~4.6, GPT-5.1, GPT-4o, GPT-4.1, DeepSeek~V3.1,
Qwen~3.5) and the heuristic baseline as a null control were tested.
Because the heuristic carries no priors, any direction effect it shows
reflects a substrate difference in how easy each direction is to learn,
not prior reliance; we therefore also report
$\Delta_{\text{model}} - \Delta_{\text{heuristic}}$ to isolate the
prior-reliance component.

\paragraph{Healthy mode --- no prior-reliance signal.}
Healthy-mode composite scores are statistically indistinguishable
across directions for six of the seven LLMs
(Table~\ref{tab:backwards-healthy}). The exception, GPT-4o, is
significant at $\alpha = 0.05$ but goes the ``wrong'' way: it scores
higher on backwards organisms. Heuristic-adjusted
deltas are small and mixed in sign, and the mixed-effects model shows
no significant model$\times$direction interaction. Healthy-mode
prediction is learned from observation, not from expectations about how
GRNs ``should'' behave.

\begin{table}[h]
\centering
\caption{Backwards-organism ablation, healthy mode. Composite score,
mean $\pm$ std over $N{=}20$ paired seeds (Qwen: $N{=}16$ due to
Together~AI errors). $\Delta$ is forward $-$ backward; CI is the
percentile-bootstrap $95\%$ interval; $p$ is the paired Wilcoxon
signed-rank statistic. The rightmost column subtracts the heuristic's
$\Delta$ to isolate the LLM-specific prior-reliance signal from the
substrate effect.}
\label{tab:backwards-healthy}
\small
\setlength{\tabcolsep}{5pt}
\begin{tabular}{lcccccc}
\toprule
Model & Forward & Backward & $\Delta$ & 95\% CI & $p$ & $\Delta - \Delta_{\text{heur}}$ \\
\midrule
Claude Opus 4.6   & 0.908 $\pm$ 0.101 & 0.933 $\pm$ 0.100 & $-0.025$ & $[-0.092, +0.042]$ & 0.50 & $-0.017$ \\
Claude Sonnet 4.6 & 0.908 $\pm$ 0.101 & 0.867 $\pm$ 0.168 & $+0.042$ & $[-0.042, +0.125]$ & 0.66 & $+0.050$ \\
GPT-5.1           & 0.750 $\pm$ 0.127 & 0.767 $\pm$ 0.166 & $-0.017$ & $[-0.100, +0.067]$ & 0.67 & $-0.008$ \\
GPT-4o            & 0.658 $\pm$ 0.138 & 0.733 $\pm$ 0.137 & $-0.075$ & $[-0.125, -0.025]$ & \textbf{0.017} & $-0.067$ \\
GPT-4.1           & 0.733 $\pm$ 0.137 & 0.708 $\pm$ 0.119 & $+0.025$ & $[-0.050, +0.108]$ & 0.56 & $+0.033$ \\
DeepSeek V3.1     & 0.925 $\pm$ 0.101 & 0.858 $\pm$ 0.146 & $+0.067$ & $[-0.008, +0.150]$ & 0.17 & $+0.075$ \\
Qwen 3.5          & 0.875 $\pm$ 0.114 & 0.885 $\pm$ 0.199 & $-0.010$ & $[-0.115, +0.104]$ & 0.72 & $-0.002$ \\
\midrule
Heuristic         & 0.583 $\pm$ 0.148 & 0.592 $\pm$ 0.175 & $-0.008$ & $[-0.083, +0.075]$ & 0.73 & --- \\
\bottomrule
\end{tabular}
\end{table}

\paragraph{Interpretation.}
Prediction under a fixed experimental budget does not depend on
real-biology priors: inverting every sign in the organism leaves
healthy-mode scores unchanged for six of the seven LLMs; the seventh
scores higher on the inverted organism, which prior reliance cannot
produce. This supports the
benchmark's central design assumption, that the prediction task is
solved from the evidence an agent collects rather than from
expectations about how GRNs ``should'' behave. It does not speak to
diagnosis, which this ablation does not test (see Limitations).

\paragraph{Limitations.}
(i)~Healthy mode and the medium tier only. The ablation therefore
constrains prediction and says nothing about whether priors affect
diagnosis; backwards-convergence issues on the hard tier (mutual
inhibition from flipped feedback edges) are deferred to future work.
(ii)~The mixed-effects fit covers the original five LLMs + heuristic;
DeepSeek and Qwen, added in a follow-up sweep, are covered only by
per-model paired tests because their session-level JSONLs are not
co-located with the original five-LLM run.
(iii)~Qwen completed on $16/20$ healthy seeds (Together~AI errors);
paired tests use the seed-intersection.
(iv)~A null result of this kind bounds a prior-reliance effect rather
than excluding one: with $N{=}20$ paired seeds the confidence intervals
still admit deltas of roughly $\pm 0.15$.

\subsection{Vocabulary and prompt-permission ablation}
\label{app:vocab}

\paragraph{Design.}
\luca{}'s invented gene vocabulary and ``biological knowledge is
irrelevant'' system-prompt assertion are design choices intended to
prevent retrieval-based shortcuts. To test whether either is
load-bearing, we ran a $2 \times 2 \times 2$ paired ablation: gene-name
\textbf{vocabulary} (invented \texttt{G001}, \dots\ vs.\ 50 real human
gene names like \texttt{TP53}, \texttt{BRCA1} drawn from three
familiarity tiers), system-prompt \textbf{stance} (\emph{suppress}: the
default ``biological knowledge is irrelevant''; \emph{permit}: ``you
may draw on biological knowledge where relevant''), and
\textbf{difficulty} (medium: $20$ genes; hard: $30$ genes). Each
organism topology was generated once and run under all four
(vocabulary $\times$ stance) conditions on Claude Opus~4.6 and DeepSeek
V3.1, $20$ seeds per cell ($320$ sessions total).

\paragraph{Results.}
Both axes are null across the $320$-session grid
(Table~\ref{tab:vocab}). Vocabulary (real vs.\ invented) shows no
measurable effect in any of $8$ paired contrasts ($|\Delta| \leq
0.058$). Stance (permit vs.\ suppress) is null in $7$ of $8$ cells; the
one nominally significant cell (DeepSeek, hard, real; $p = 0.035$)
fails Bonferroni correction across the $8$ tests and points the
``wrong'' way --- permit \emph{decreases} performance.

\begin{table}[h]
\centering
\caption{Vocabulary and prompt-permission ablation (medium and hard
healthy, $N{=}20$ seeds per cell, $320$ sessions).
\textbf{inv}~=~invented vocabulary, \textbf{real}~=~real human gene
names; \textbf{sup}~=~suppress biology, \textbf{per}~=~permit biology.
Mean composite score $\pm$ std.}
\label{tab:vocab}
\small
\begin{tabular}{llcccc}
\toprule
Tier & Model & inv / sup & real / sup & inv / per & real / per \\
\midrule
Medium & Claude Opus 4.6   & 0.917 $\pm$ 0.101 & 0.908 $\pm$ 0.114 & 0.925 $\pm$ 0.114 & 0.917 $\pm$ 0.101 \\
Medium & DeepSeek V3.1     & 0.892 $\pm$ 0.098 & 0.842 $\pm$ 0.175 & 0.853 $\pm$ 0.209 & 0.825 $\pm$ 0.262 \\
Hard   & Claude Opus 4.6   & 0.592 $\pm$ 0.206 & 0.608 $\pm$ 0.146 & 0.650 $\pm$ 0.187 & 0.608 $\pm$ 0.182 \\
Hard   & DeepSeek V3.1     & 0.567 $\pm$ 0.183 & 0.608 $\pm$ 0.165 & 0.533 $\pm$ 0.220 & 0.475 $\pm$ 0.182 \\
\bottomrule
\end{tabular}
\end{table}

\paragraph{Interpretation.}
\luca{} organisms are generated with random module assignment and random
gene labelling, so biological priors cannot map onto the causal
structure an agent must recover. Performance reflects experiment-driven
reasoning regardless of whether gene identifiers are real or invented
and regardless of whether the prompt invites biological reasoning. The
invented-vocabulary design choice turns out not to be load-bearing on
its own --- but the null result here is what makes the benchmark's
retrieval-free claim empirical rather than asserted. This ablation
addresses an LLM-era concern that does not arise for classical
synthetic-GRN benchmarks~\citep{vandenbulcke2006syntren,
schaffter2011genenetweaver, tjarnberg2017genespider}, which evaluate
inference algorithms on static observational data with no training
corpus to memorise.

\paragraph{Coverage and limitations.}
The ablation covers two model families (Anthropic and Together-hosted
DeepSeek) at two difficulties; broader coverage (Sonnet~4.6, GPT-5.1)
remains future work but would test cross-provider generalisation rather
than the headline null. We note that both axes were tested only on
healthy mode; disease-mode results would require redoing the ablation
with malfunction-targets randomised under each vocabulary condition.

\clearpage

\end{document}